# Bioacoustic Signal Classification Based on Continuous Region Processing, Grid Masking and Artificial Neural Network

**Mohammad Pourhomayoun**  MP749@CORNELL.EDU
**Peter J. Dugan**  PJD78@CORNELL.EDU
**Marian Popescu**  CP478@CORNELL.EDU
**Christopher W. Clark**  CWC2@CORNELL.EDU
Bioacoustics Research Program (BRP), Cornell University, Ithaca, NY, USA, 14850

### Abstract

In this paper, we develop a novel method based on machine-learning and image processing to identify North Atlantic right whale (NARW) up-calls in the presence of high levels of ambient and interfering noise. We apply a continuous region algorithm on the spectrogram to extract the regions of interest, and then use grid masking techniques to generate a small feature set that is then used in an artificial neural network classifier to identify the NARW up-calls. It is shown that the proposed technique is effective in detecting and capturing even very faint up-calls, in the presence of ambient and interfering noises. The method is evaluated on a dataset recorded in Massachusetts Bay, United States. The dataset includes 20000 sound clips for training, and 10000 sound clips for testing. The results show that the proposed technique can achieve an error rate of less than FPR = 4.5% for a 90% true positive rate.

## 1. Introduction and Background

Bioacoustic signal detection and classification is one of the most common and effective techniques used by scientists to explore marine bioacoustics and understand marine mammal behavioral patterns. For passive marine acoustic research, hydrophone sensor systems collect huge amounts of underwater sound data, thereby placing a premium on automated computer algorithms, including machine-learning methods, for detecting and classifying sounds of interest in the data (Sousa-Lima et al. 2013).

Right whales produce frequency-modulated upsweeps, referred to as up-calls, for long-range communication in the 50-250Hz frequency band (Clark 1982), and detection of up-calls has been shown to be the most effective mechanisms for determining whale presence in critical habitats (Clark et al. 2010). In this paper, we develop novel methods based on image processing and machine-learning to detect North Atlantic Right Whales (NARW) up-calls in the presence of high environmental noise, using a fairly small feature set (5, 15 or 20 features). The NARW is one of the world's most highly endangered whales (Clapham *et al.* 1999). Therefore, there is an urgent need to develop efficient techniques to detect the presence of NARWs so as to determine their seasonal occurrences and protect them from possible harm (Kraus et al. 2005).

For decades, researchers have been working to design effective automated algorithms for identifying marine mammal vocalizations including NARW up-calls (Mellinger 2004; Mohammad *et al.* 2011). Mellinger compared different methods for up-call detection, including spectrogram correlation and an artificial neural network. He evaluated the spectrogram correlation method for two different cases; manually selected parameters, and parameters selected based on an optimization procedure. In another approach, he used spectrogram frames, consisting of 252 cells, as inputs to a feed-forward neural network with 10 hidden layers. He used standard gradient-descent back-propagation with 5000 epochs to train the neural network (Mellinger 2004). However, the size of the selected neural network and the large number of inputs lead to extremely high computational complexity and a long training time. Dugan *et al.* also developed two new approaches for NARW sound identification based on artificial neural networks and decision tree classifiers, and compared their performance to a multi-stage feature vector testing (FVT) method (Dugan *et al.* 2010).

Gillespie applied an edge detection algorithm on the smoothed spectrogram to determine the boundary of the sound. He then extracted features, including duration, bandwidth and details of the frequency contour, which were used in an up-call classification stage (Gillespie 2004). Sánchez-García *et al.* also used a spectrogram region-based segmentation technique to identify the sound signal, and then extracted the mean values of a fixed number of radial basis function (RBF) coefficients. These coefficients were later used to classify the sound signals (Sánchez-García *et al.* 2009). Mohammad *et al.* (2011) developed a region-based active contour model and support vector machine classifier to identify the NARW up-call in shallow water.

It is important to note that in environments with high ambient noise levels and various amounts of acoustic clutter, including sounds from other species, NARW up-calls can be extremely difficult to detect. Thus, regular region growing techniques, or methods based on using the maximum spectrogram value as the initial point of contour segmentation usually fail to find the up-call in cases of low SNR or high clutter.

In this paper, we develop a new method, based on continuous region processing and grid masking methods, to detect NARW up-calls in the presence of ambient noise, interfering noise or other non-NARW sounds. We apply a continuous region algorithm on the de-noised and normalized spectrogram to extract continuous regions of interest that might represent portions of an up-call. Then, we use grid masking techniques to generate two sets of features that are used as inputs to an artificial neural network classifier to identify the NARW up-calls.

## 2. Methods

In this section, we describe the details of the proposed method for up-call identification. Figure 1 shows the block diagram and different steps of the proposed approach.

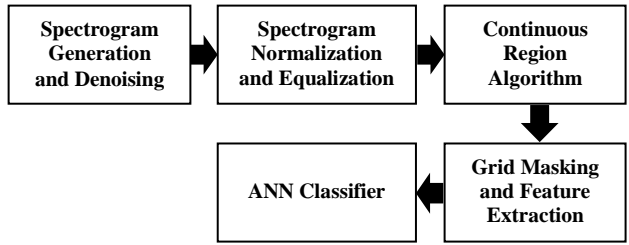

Figure 1. Block diagram of the proposed method.

### 2.1 Spectrogram Normalization and Equalization

The sound signals, sampled at 2 kHz, are clipped into 2 sec slices, which are used to generate time-frequency spectrograms. To produce the spectrograms, we apply a STFT with window size of 128 ms (Hann window, 256 samples, 50% overlap).

After producing a spectrogram, we apply a two-dimensional wiener filter in order to denoise and smooth the spectrogram, using a 5x5 window around each pixel to estimate the local variance (Lim 1990). For each frequency band, we zero-mean the denoised spectrogram to remove the effects of constant narrowband noise, such as ship tonals, wind noise or electrical device noise, and to emphasize short-duration FM sounds such as NARW up-calls (Mellinger 2004).

The next step is hard-limiting the upper and lower bounds of spectrogram amplitudes to remove the influence of extreme values (Mellinger 2004):

$$\hat{S}(t,f) = \max(S_{floor}, \min(S_{ceiling}, S_N(t,f))) - S_{floor} \quad (1)$$

where $S_N(t,f)$ is the normalized spectrogram, and $S_{floor}$ and $S_{ceiling}$ are the desired lower and upper bounds on spectrogram values. Figure 2 shows the effect of denoising, normalization and equalization on a sample spectrogram.

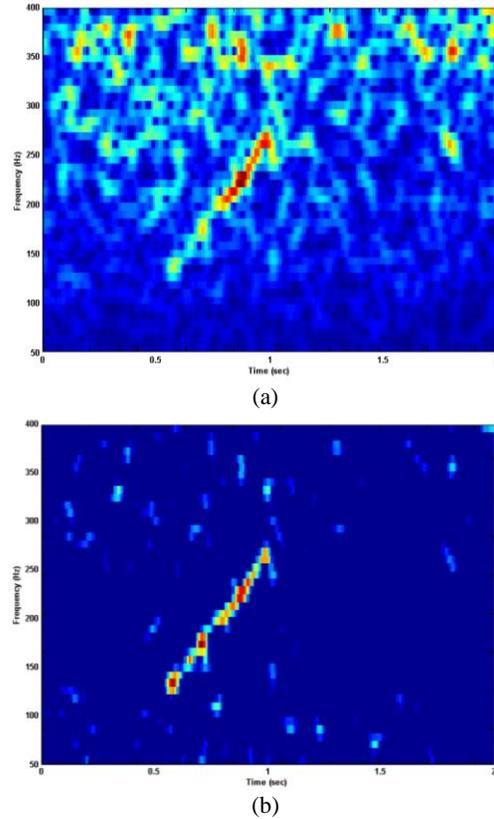

Figure 2. Spectrogram examples with and without denoising, normalization and equalization: (a) original spectrogram. (b) denoised, normalized and equalized spectrogram.

### 2.2 Continuous Region Processing

After the denoising, normalization and equalization steps, we convert the spectrogram into a binary image for the purpose of continuous region processing. Note that this binary image will only be used to find regions in the spectrogram that are considered part of an up-call; however, the output of the algorithm will be a non-binary spectrogram including the regions of interest. Since some of the up-calls are extremely faint, we set the threshold value very low (e.g 10% of the image mean) to reduce the chances that we would miss an object of interest.

We use the Moore-Neighbor tracing algorithm modified by Jacob's stopping criteria (Gonzalez, 2004) to determine objects (i.e. continuous regions) in the image. After that, we extract the properties of each object and compare them to a set of thresholds to find an up-call. As mentioned before, under conditions of low SNR, NARW up-calls can be completely buried in noise, and typical region-growing techniques fail to find the up-call. Furthermore, as shown in Figure 5, sometimes up-calls are extremely faint compared to other objects in the spectrogram (e.g. interfering noise or sounds of other species). In such situations, methods that use the maximum spectrogram value to identify an initial point contour segmentation are not able to detect and classify the up-call.

Table 1 shows the continuous region parameters and thresholds used for detecting up-call segments in spectrograms given the specifications mentioned in 2.1. Figures 3 and 4 illustrate the continuous region algorithm process, and the elimination of noise and other possible sound objects in the frame that do not meet the NARW up-call criteria. The spectrogram in Figure 3 (top panel) includes high ambient noise, while the spectrogram in Figure 4 (top panel) contains other sound objects in the same frequency band.

Figure 5 demonstrates how the proposed technique performs under very challenging conditions when the NARW up-call is very faint and SNR is very poor.

*Table1.* The continuous region parameters and the thresholds used for detecting and north right whale up-call piece.

| Parameter | Threshold |
|---|---|
| Minimum Perimeter | 15 pixel |
| Minimum Area | 15 pixel |
| Minimum Height (frequency band) | 14 Hz |
| Maximum Height (frequency band) | 250 Hz |
| Minimum Width (duration) | 0.1 sec |
| Maximum Width (duration) | 2 sec |
| Minimum orientation of the surrounding Ellipse | 1° |
| Maximum orientation of the surrounding Ellipse | 88° |
| Minimum Height/Width ratio | 0.05 |
| Maximum Height/Width ratio | 3 |
| Minimum Frequency | 50 Hz |
| Maximum Frequency | 400 Hz |
| Maximum surrounding Ellipse axes ratio | 3.5 |

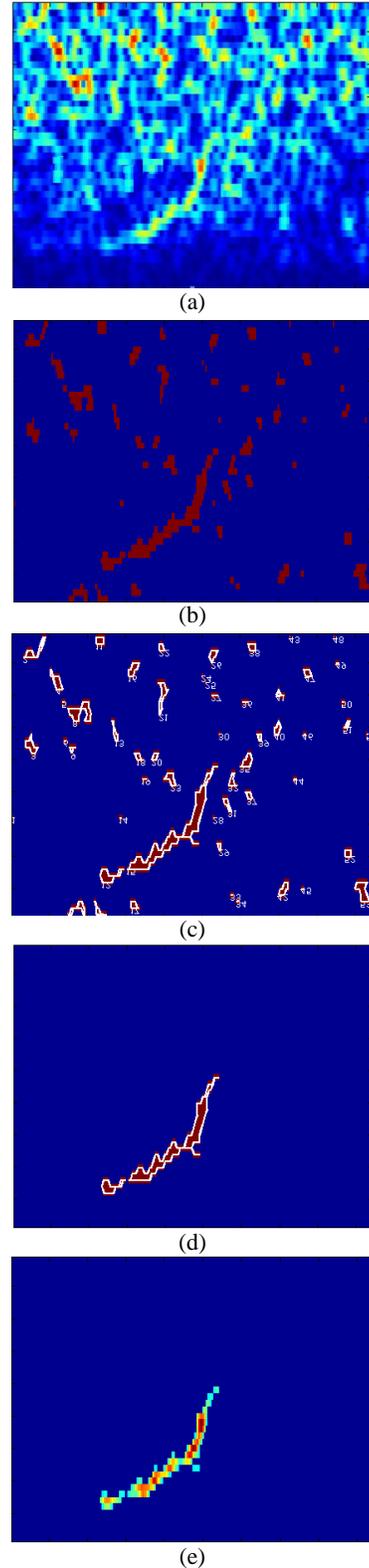

Figure 3. Continuous Region Processing: (a) original Spectrogram; (b) spectrogram after denoising, normalization, equalization and binarization; (c) continuous region detection; (d) detected region of interest; and (e) the algorithm's output.

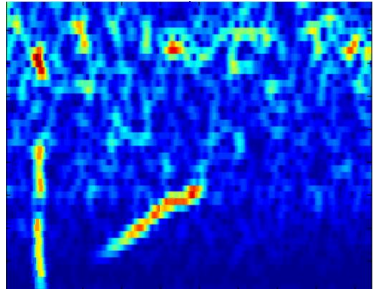

(a)

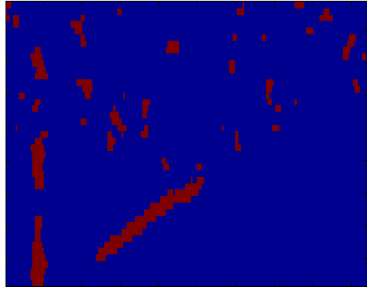

(b)

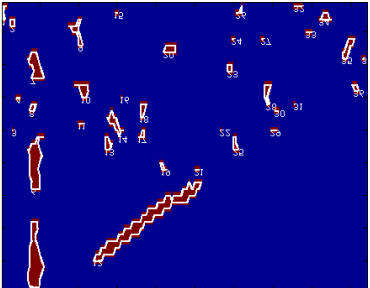

(c)

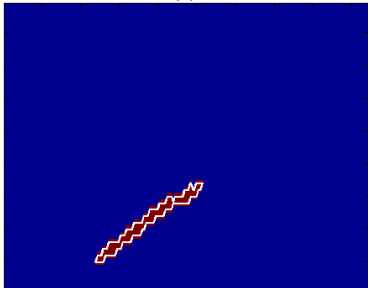

(d)

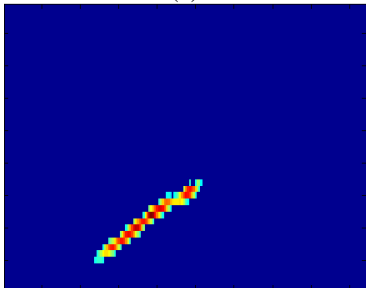

(e)

Figure 4. Continuous Region Processing: (a) original Spectrogram; (b) Spectrogram after denoising, normalization, equalization and binarization; (c) continuous region detection, (d) detected region of interest; and (e) the algorithm output.

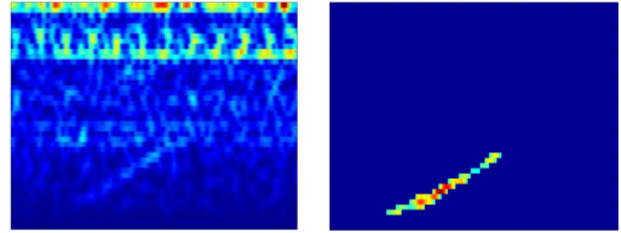

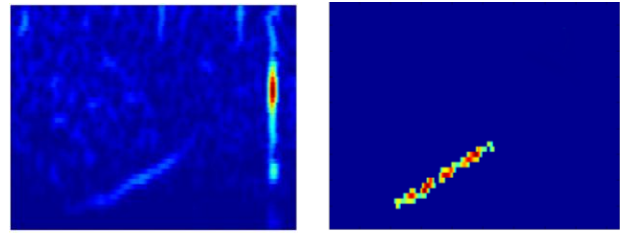

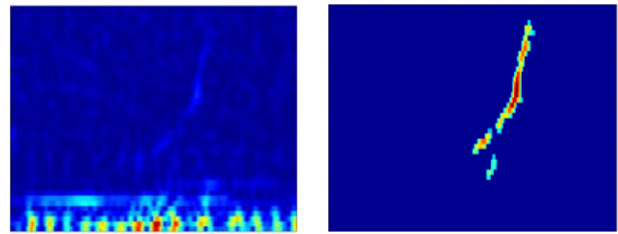

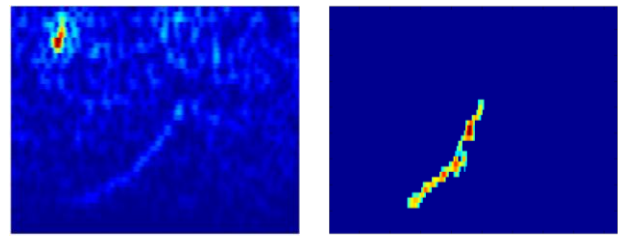

Figure 5. Four examples of faint NARW up-calls and outputs from the continuous region algorithm process. Left: Original spectrogram. Right: Proposed continuous region algorithm output.

### 2.3 Grid Masking and Feature Extraction

After continuous region processing and generation of the new spectrogram including the regions of interest, we divide the new spectrogram into equally spaced grids. As shown in Figure 6-(a), we used a 6x6 grid for the spectrogram with the specifications mentioned in 2.1.

The first set of features includes the means of spectrogram values over minor diagonals of the grid plane. Figure 6-(b) shows the grid pattern used to extract the diagonal features. In this figure, the diagonal grid cells are distinguished using colors and numbers. For example, in Figure 6-(a), the grid cells over diagonal #3 have a significant mean value compared to other diagonals.

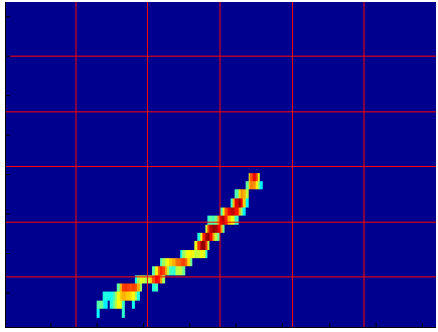

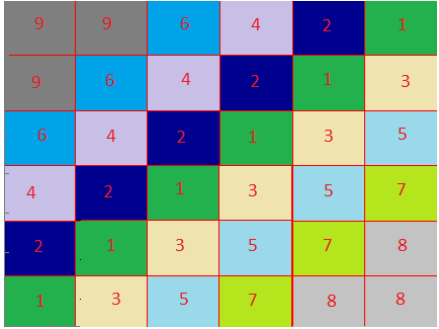

Figure 6. Spectrogram gridding: (a) sample spectrogram gridding, (b) the grid pattern used to extract the diagonal features. In this case (6x6 gridding), the diagonal feature set includes 9 features corresponding to the means of spectrogram values over the diagonal grid cells shown above.

The second set of features is generated using the sliding masks shown in Figure 7. These are binary masks having the value of one for black cells and zero for white cells. Each mask slides over the grid plane and calculates the averages of the spectrogram mean values located in black cells. The feature for each grid cell is determined as the maximum value of the three masking results as following,

$$f(x, y) = \max\{M_1(x, y), M_2(x, y), M_3(x, y)\} \quad (2)$$

$$M_1(x, y) = (mean(x+1, y) + mean(x, y+1) + mean(x+1, y+1))/3$$
$$M_2(x, y) = (mean(x+1, y) + mean(x, y+1))/2 \quad (3)$$
$$M_3(x, y) = (mean(x, y) + mean(x, y+1))/2$$

where $f(x,y)$ is the feature allocated to the each grid cell $(x,y)$; $M_1(x,y)$, $M_2(x,y)$, and $M_3(x,y)$ are the masking results for grid cell $(x,y)$ corresponding to the three masks shown in Figure 7; and $mean(x,y)$ is the mean value of spectrogram points located inside the grid cell $(x,y)$.

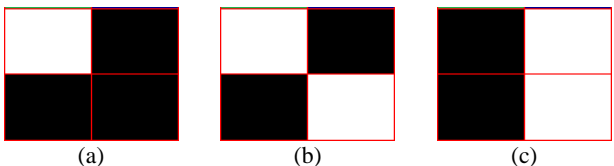

(a)      (b)      (c)

Figure 7. The three masks used to extract the grid features

## 2.4 Artificial Neural Network

The Artificial Neural Network (ANN) is a popular and effective technique for bioacoustic signal classification (Potter et al. 1994; Mellinger 2004). ANNs can accept a wide range of feature variables as input, such as those from a spectrogram, frequency contour, or waveform. This flexibility allows the application of ANNs to different detection conditions and various types of signals. For example, Potter *et al.* used a feed-forward ANN to distinguish bowhead whale endnotes from interfering noises. They used the signal spectrogram as the input to the ANN (Potter *et al.* 1994). In another example, Deecke *et al.* used a standard back-propagation trained ANN to classify killer whale dialects to nine different categories (Deecke *et al.* 1999). They used the extracted pulse-rate contours as the input to the ANN.

There are several types of ANNs that can be applied for the purpose of classification. Feed-forward networks are more preferred for our purpose because they have less complexity, and relatively fewer numbers of neurons and connections compared to feedback networks (Potter, et al. 1994). However, feed-forward networks usually need a larger training dataset for backpropagation training (Potter, et al. 1994). In our case, this is not a problem because we have a large enough dataset available for training purpose. Therefore, we chose a standard, feed-forward, back-propagation trained network for classification. In this problem, we use a network with two hidden layers that receives the feature vectors extracted from the spectrograms (as described in 2.1, 2.2 and 2.3) as input.

## 3. Results and Conclusion

The proposed method was evaluated on a dataset recorded in Massachusetts Bay, United States. The dataset includes 20000 sound clips for training (containing 4473 NARW up-calls, and 15527 non-up-calls), and 10000 sound clips for testing (containing 2554 NARW up-calls, and 7446 non-up-calls). After spectrogram denoising, normalization, and equalization, we applied the proposed continuous region processing to detect the regions of interest, and extract the features as presented in 2.3. In this case, we extracted and used only 20 features including the 5 diagonal features corresponding to diagonals 1-5 in Figure 6-b, and 15 masking features generated by sliding the masks shown in Figure 7 over the spectrogram except for the first and last columns. We used an ANN classifier with only 2 hidden layers, with sizes of 32 and 16 neurons, trained using standard gradient-descent back-propagation with 100 epochs. The proposed method was evaluated for three different cases: with only 5 diagonal features as the feature set, with only 15 masking features as the feature set, and with the total 20 features as the feature set for ANN training, testing and classification. Figure 8 demonstrates the Receiver Operating Curve (ROC) of the proposed method evaluated on the testing dataset for the three different cases. Table 2 also shows the error rate (False Positive) for a fixed 90% True Positive Rate (e.g FPR=4.5% for TPR=90% using all 20 features). As we see in Figure 8 and Table 2, the proposed method can achieve high performance even using fewer numbers of features (5 features). This can be very beneficial when we aim to reduce the computational complexity of the classification stage.

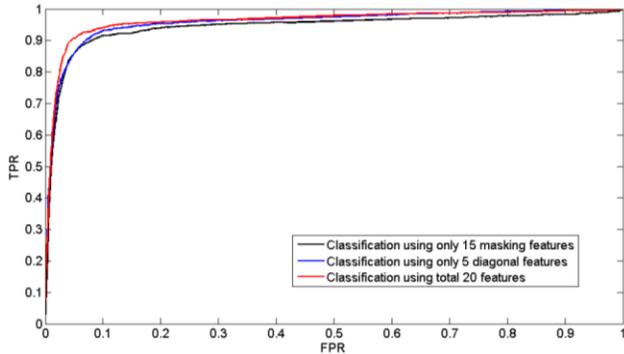

Figure 8. Receiver Operating Curve (ROC) of proposed method for the testing dataset

*Table 2.* The performance of the proposed method for fixed 90% True Positive Rate using diagonal and masking features.

| Features | FPR% | TPR% |
|---|---|---|
| Using total 20 features | 4.5 | 90 |
| Using only 5 diagonal features | 7.0 | 90 |
| Using only 15 masking features | 8.1 | 90 |

As shown in Figures 3 and 4, the proposed technique is effective for detecting up-calls under conditions of high ambient noise and/or interfering sounds. Figure 5 also illustrates that this method is also capable of capturing even very faint up-calls (i.e. low SNRs). Furthermore, this approach performs very well despite the relatively low number of features (Table 2). Future directions for this work include applying this algorithm to large continuous archival sounds streams and investigating the performance for recognizing NARW calls within context of accurate seasonal information.